\documentclass{article}
\usepackage{ijcai18}

\usepackage{times}
\usepackage{xcolor}
\usepackage{soul}
\usepackage[utf8]{inputenc}
\usepackage[small]{caption}

\usepackage[pdftex]{graphicx}

\usepackage[hidelinks]{hyperref} 
\usepackage{algorithmicx, algorithm, algpseudocode}
\usepackage{amsmath, amssymb}
\usepackage{xspace}
\usepackage{caption}
\usepackage{subcaption}
\usepackage{wrapfig}
\usepackage{tabu}
\usepackage{color}

\newcommand{\EZL}{{\em EZLearn}\xspace}

\newcommand{\smallpar}[1]{\vspace{-12.5pt}\paragraph{#1}}

\begin{document}
\title{\EZL: Exploiting Organic Supervision in Automated Data Annotation}

\author{
Maxim Grechkin$^1$, 
Hoifung Poon$^2$, 
Bill Howe$^1$
\\ 
$^1$ University of Washington \\
$^2$ Microsoft Research\\
grechkin@uw.edu,
hoifung@microsoft.com,
billhowe@uw.edu
}

\maketitle
\begin{abstract}

Many real-world applications require automated data annotation, such as identifying tissue origins based on gene expressions and classifying images into semantic categories.
Annotation classes are often numerous and subject to changes over time, and annotating examples has become the major bottleneck for supervised learning methods.
In science and other high-value domains, large repositories of data samples are often available, together with two sources of {\em organic supervision}: a lexicon for the annotation classes, and text descriptions that accompany some data samples.
Distant supervision has emerged as a promising paradigm for exploiting such indirect supervision by automatically annotating examples where the text description contains a class mention in the lexicon.
However, due to linguistic variations and ambiguities, such training data is inherently noisy, which limits the accuracy of this approach.
In this paper, we introduce an auxiliary natural language processing system for the text modality, and incorporate co-training to reduce noise and augment signal in distant supervision.
Without using any manually labeled data, our \EZL system learned to accurately annotate data samples in functional genomics and scientific figure comprehension, substantially outperforming state-of-the-art supervised methods trained on tens of thousands of annotated examples.

\end{abstract}

\section{Introduction}

The confluence of technological advances and the open data movement \cite{molloy2011open} has led to an explosion of publicly available datasets, heralding an era of data-driven hypothesis generation and discovery in high-value applications \cite{Piwowar2013}.
A prime example is {\em open science}, which promotes open access to scientific discourse and data to facilitate broad data reuse and scientific collaboration \cite{Friesike2015}.
In addition to enabling reproducibility, this trend has the potential to accelerate scientific discovery, reduce the cost of research, and facilitate automation \cite{rung2013reuse,libbrecht2015machine}.

However, progress is hindered by the lack of consistent and high-quality annotations. 
For example, 
the NCBI Gene Expression Omnibus (GEO) \cite{Clough2016} contains over two million gene expression profiles, yet only a fraction of them have explicit annotations indicating the tissue from which the sample was drawn, information that is crucial to understanding cell differentiation and cancer \cite{hanahan2011hallmarks,gutierrez2015tissue}. As a result, only 20\% of the datasets have ever been reused, and tissue-specific studies are still only performed at small scales \cite{Piwowar2013}.

Annotating data samples with standardized classes is the canonical multi-class classification problem, but standard supervised approaches are difficult to apply in these settings. Hiring experts to annotate examples for thousands of classes such as tissue types is unsustainable. Crowd-sourcing is generally not applicable, as annotation requires domain expertise that most crowd workers do not possess.
Moreover, the annotation standard is often revised over time, incurring additional cost for labeling new examples.

While labeled data is expensive and difficult to create at scale, unlabeled data is usually in abundant supply. Many methods have been proposed to exploit it, but they typically still require labeled examples to initiate the process \cite{blum1998combining,mcclosky2008self,fei2006one}. 
Even zero-shot learning, where the name implies learning with no labeled examples for {\em some} classes, still requires labeled examples for related classes \cite{palatucci2009zero,socher2013zero}.

In this paper, we propose \EZL, which makes annotation learning easy by exploiting two sources of {\em organic supervision}.
First, the annotation classes generally come with a lexicon for standardized references 
(e.g., ``liver'', ``kidney'', ``acute myeloid leukemia cell'' for tissue types).
While labeling individual data samples is expensive and time-consuming, it takes little effort for a domain expert to provide a few example terms for each class.
In fact, in sciences and other high-value applications, such a lexicon is often available from an existing ontology. For example, the Brenda Tissue Ontology specifies 4931 human tissue types, each with a list of standard names \cite{Gremse01012011}.
Second, data samples are often accompanied by a free-text description, some of which directly or indirectly mention the relevant classes (e.g., the caption of a figure, or the description for a gene expression sample). 
Together with the lexicon, these descriptions present an opportunity for exploiting distant supervision by generating (noisy) labeled examples at scale \cite{mintz2009distant}.
We call such indirect supervision ``organic'' to emphasize that it is readily available as an integral part of a given domain.

In practice, however, there are serious challenges to enact this learning process.
Descriptions are created for general human consumption, not as high-quality
machine-readable annotations. They are provided voluntarily by data owners and lack consistency; ambiguity, typos, abbreviations, and non-standard references are common \cite{Lee2013,rung2013reuse}. Multiple samples may share a text description that mentions several classes, introducing uncertainty as to which class label is associated with which sample. Additionally, annotation standards evolve over time, introducing new terms and evicting old ones. As a result, while there are potentially many data samples whose descriptions contain class information, only a fraction of them can be correctly labeled using distant supervision. This problem is particularly acute for domains with numerous classes and frequent updates, such as the life sciences.

\begin{figure}
    \centering
    \includegraphics[width=0.9\linewidth]{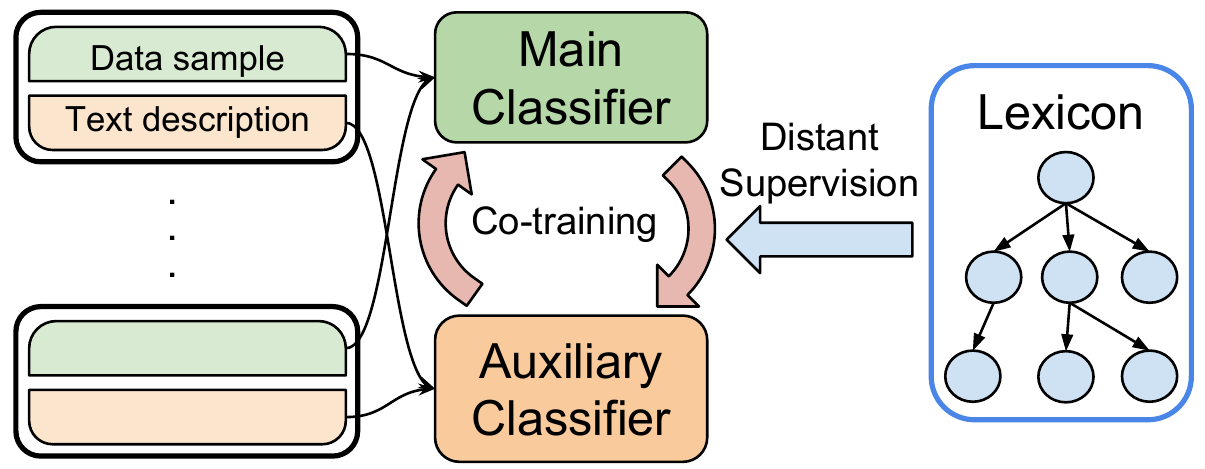}
    \caption{The \EZL architecture: an auxiliary text-based classifier is introduced to bootstrap from the lexicon (often available from an ontology) and co-teaches the main classifier until convergence.
                    }
    \label{fig:EZLdiagram}
\end{figure}

To best exploit indirect supervision using all instances, \EZL introduces an auxiliary text classifier for handling complex linguistic phenomena.
This auxiliary classifier first uses the lexicon to find exact matches to teach the main classifier. In turn, the main classifier helps the auxiliary classifier improve by annotating additional examples with non-standard text mentions and correcting errors stemming from ambiguous mentions. This co-supervision continues until convergence. Effectively, \EZL represents the first attempt in combining distant supervision and co-training, using text as the auxiliary modality for learning (Figure~\ref{fig:EZLdiagram}).

To investigate the effectiveness and generality of \EZL, we applied it to two important applications:  functional genomics and scientific figure comprehension, which differ substantially in sample input dimension and description length.
In functional genomics, there are thousands of relevant classes. In scientific figure comprehension, prior work only considers three coarse classes, which we expand to twenty-four. In both scenarios, \EZL successfully learned an accurate classifier with zero manually labeled examples. 

While standard co-training has labeled examples from the beginning, \EZL can only rely on distant supervision, which is inherently noisy. 
We investigate several ways to reconcile distant supervision with the trained classifier's predictions during co-training. 
We found that it generally helps to ``remember'' distant supervision while leaving room for correction, especially by accounting for the hierarchical relations among classes.
We also conducted experiments to evaluate the impact of noise on \EZL. The results show that \EZL can withstand a large amount of simulated noise without suffering substantial loss in annotation accuracy.
 \vspace{-5pt}
\section{Related Work}

A perennial challenge in machine learning is to transcend the supervised paradigm by making use of unlabeled data. 
Standard unsupervised learning methods cluster data samples by explicitly or implicitly modeling similarity between them. It cannot be used directly for classification, as there is no direct relation between learned clusters and  annotation classes.

In semi-supervised learning, direct supervision is augmented by annotating unlabeled examples using either a learned model \cite{nigam2000analyzing,blum1998combining} or similarity between examples \cite{zhu2002learning}.
It is an effective paradigm to refine learned models, but still requires initialization with sufficient labeled examples for all classes. 
Zero-shot learning or few-shot learning relax the requirement of labeled examples for some classes, but still need to have sufficient labeled examples for \emph{related} classes \cite{palatucci2009zero,socher2013zero}. In this regard, they bear resemblance with domain adaptation \cite{blitzer2007biographies,daume2009frustratingly} and transfer learning \cite{pan2010survey,raina2007self}. 
Zero-shot learning also faces additional challenges such as novelty detection to distinguish between known classes and new ones.

An alternative approach is to ask domain experts to provide example annotation functions, ranging from regular expressions \cite{hearst1991noun} to general programs \cite{ratner2016data}. Common challenges include combating low recall and semantic drifts. 
 Moreover, producing useful annotation functions still requires domain expertise and substantial manual effort, and may be impossible when predictions depend on complex input patterns (e.g., gene expression profiles).

\EZL leverages domain lexicons to annotate noisy examples from text, similar to distant supervision \cite{mintz2009distant}. However, distant supervision is predominantly used in information extraction, which considers the single view on text \cite{quirk2016distant,peng2017cross}.
In \EZL, the text view is introduced to support the main annotation task, resembling co-training \cite{blum1998combining}. The original co-training algorithm annotates unlabeled examples in batches, where \EZL relabels all examples in each iteration, similar to co-EM \cite{nigam2000analyzing}.

\vspace{-5pt}
\section{\EZL}

Let $X=\{x_i: i\}$ be the set of data samples and $C$ be the set of classes.
Automating data annotation involves learning a multi-class classifier $f: X\rightarrow C$.
For example, $x_i$ may be a vector of gene expression measurements for an individual, where $C$ is the set of tissue types.
Additionally, we denote $t_i$ as the text description that accompanies $x_i$. 
If the description is not available, $t_i$ is the empty string.

\begin{algorithm}[t]
\begin{algorithmic}
\caption{\EZL}\label{EZL-algorithm}
\State \textbf{Input:} Data samples $X$, text descriptions $T$, annotation classes $C$, and lexicon $L$ containing example terms $L_c$ for each class $c\in C$.
\State \textbf{Output:} Trained classifiers $f: X\rightarrow C$ (main) and $f_T: T\rightarrow C$ (auxiliary).  \State \textbf{Initialize:} Generate initial training data $D^0$ as all $(x_i, t_i, c)$ for $x_i\in X$, $t_i\in T$, where $t_i$ mentions some term in $L_c$.

\For{\texttt{$k=1:N_{iter}$}}
    \State $f \leftarrow \mathit{\tt Train_{main}}(D^{k-1})$; $D_T^k \leftarrow {\tt Resolve}(f(X), D^0)$
    
    \State $f_T \leftarrow \mathit{\tt Train_{aux}}(D_T^{k})$; $D^k \leftarrow {\tt Resolve}(f_T(T), D^0)$
  
\EndFor
\end{algorithmic}
\end{algorithm}  

Algorithm~\ref{EZL-algorithm} shows the \EZL algorithm.
By default, there are no available labeled examples $(x, y^*)$ where $y^*\in C$ is the true class for annotating $x\in X$. 
Instead, \EZL assumes that a lexicon $L$ is available with a set of \emph{example terms} $L_c$ for each $c \in C$.
We do not assume that $L_c$ contains every possible synonym for $c$, nor that such terms are unambiguous. Rather, we simply require that $L_c$ is non-empty for any $c$ of interest.
We use $L_c$'s for distant supervision in \EZL, by creating an initial labeled set $D^0$, which consists of all $(x_i, t_i, c)$ where the text description $t_i$ explicitly contains at least one term in $L_c$.

To handle linguistic variations and ambiguities, \EZL introduces an auxiliary classifier $f_T: T\rightarrow C$, where $T=\{t_i: i\}$. At iteration $k$, we first train a new main classifier $f^k$ using $D^{k-1}$. We then apply $f^k$ to $X$ and create a new labeled set $D_T^k$, which contains all $(t_i, c)$ where $f^k(x_i)=c$.
We then train a new text classifier $f_T^k$ using $D_T^k$, and create the new labeled set $D^k$ with all $(x_i,c)$ where $f_T^k(t_i)=c$. This process continues until convergence, which is guaranteed given  independence of the two views conditioned on the class label \cite{blum1998combining}. Empirically, it converges quickly.

We can use any classifier for $\tt Train_{main}$ and $\tt Train_{aux}$. 
%Features for the main classifier are domain-specific and can be what any supervised approach might use. 
%For the text classifier, we use standard $n$-gram features, which are effective in both applications we evaluated, though it is possible to tailor them for specific domains. 
Typically, the classifiers will take a parametric form (e.g., $f(x)=f(x,\theta)$) and training with a labeled set $D$ amounts to minimize some loss function $L$ (i.e., $\theta^*=\arg\min_{\theta}~\sum_{(x,y^*)\in D}~L(f(x, \theta), y^*)$).
In this paper, we opted for simple, standard choices. In particular, we used logistic regression and fastText \cite{fastText} as the main and auxiliary classifiers, respectively.

Generally, a classifier will output a score for each class rather than predicting a single class. The score reflects the confidence in predicting the given class. \EZL generates the labeled set by adding all (sample, class) pairs for which the score crosses a hyperparameter threshold. We used 0.3 in this paper, which allows up to 3 classes to be assigned to a sample. The performance of \EZL was not sensitive to this parameter: values in (0.2, 0.6) yielded similar results. In all iterations, a labeled set might contain more than one class for a sample, which 
is not a problem for the learning algorithm and 
is useful when there is uncertainty about the correct class.

For samples with distant-supervision labels, a classifier (main or auxiliary) might predict different labels in an iteration. Since distant supervision is noisy, reconciling it with the classifier’s prediction could help correct its errors. The ${\tt Resolve}(\cdot)$ function is introduced for this purpose.
The direct analog of standard co-training returns distant-supervision labels if they are available ($\tt Standard$).
Conversely, $\tt Resolve$ could ignore distant supervision and always return the classifier's prediction ($\tt Predict$).
Alternatively, $\tt Resolve$ may return all labels ($\tt Union$) or the common ones ($\tt Intersect$).

However, none of the above approaches consider the hierarchical relations among the label classes. Suppose that the text mentions both $\tt neuron$ and $\tt leukemia$, whereas the classifier predicts $\tt leukocyte$ with high confidence. Our confidence in $\tt leukemia$ being the correct label should increase since $\tt leukemia$ is a subtype of $\tt leukocyte$, and our confidence in $\tt neuron$ should decrease.
We thus propose a more sophisticated variant of $\tt Resolve$ that captures such reasoning ($\tt Relation$).
Let $c_1, c_2$ be the two labels from distant supervision and classifier prediction, respectively. If $c_1$ and $c_2$ are the same, $\tt Relation$ returns $c=c_1=c_2$. If they have a hierarchical relation, $\tt Relation$ will return the more specific one (i.e., the subtype). Otherwise, $\tt Relation$ returns none. If distant supervision or the classifier prediction assigns multiple labels to a sample, $\tt Relation$ will return results from all label pairs. (In domains with no hierarchical relations among the classes, $\tt Relation$ is the same as $\tt Intersect$.)

\begin{table*}[!ht]
\centering
\begin{tabular}{|l|r|r|c|c|c|c|c|c|}
\hline
\textbf{Method} & \textbf{\# Labeled} & \textbf{\# All} & \textbf{AUPRC}   & \textbf{Prec@0.5}  & \textbf{Use Expression}  & \textbf{Use Text}  & \textbf{Use Lexicon} & \textbf{Use EM} \\ \hline
URSA            & $14510$ &  0 & 0.40          & 0.52    & yes & no & no & no               \\ \hline
Co-EM           & $14510$ & $116895$ & $0.51 $ & $0.61$  &  yes & yes & no & yes          \\ \hline
Dist. Sup.    & 0  & $116895$ & $0.59$ & $0.63$ & yes & yes & yes & no \\ \hline
\EZL            & 0  & $116895$ & $\mathbf{0.69}$ & $\mathbf{0.86}$ & yes & yes & yes & yes \\ \hline
\end{tabular}
\vspace{-4pt}
\caption{Comparison of test results between \EZL and state-of-the-art supervised, semi-supervised, and distantly supervised methods on the CMHGP dataset.
We reported the area under the precision-recall curve (AUPRC) and precision at 0.5 recall. 
\EZL requires no manually labeled data, and substantially outperforms all other methods. 
Compared to URSA and co-EM, \EZL can effectively leverage unlabeled data by exploiting organic supervision from text descriptions and lexicon.
\EZL amounts to initializing with distant supervision (first iteration) and continuing with an EM-like process as in co-training and co-EM, which leads to further significant gains. }
\label{fig:genomicsTable}
\vspace{-6pt}
\end{table*}

\vspace{-5pt}
\section{Application: Functional Genomics}

\begin{figure}
    \centering
    \includegraphics[width=0.8\linewidth]{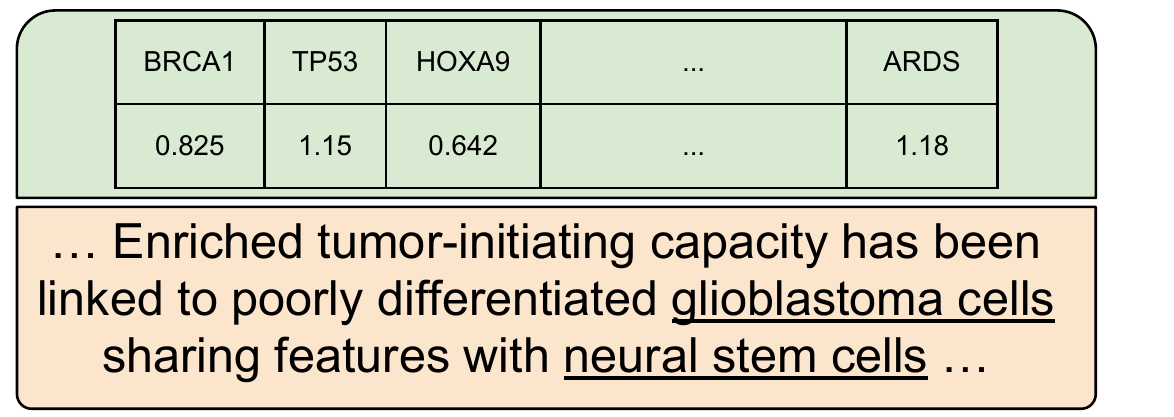}
    \vspace{-4pt}
    \caption{Example gene expression profile and its text description in Gene Expression Omnibus (GEO). Description is provided voluntarily and may contain ambiguous or incomplete class information.}
    \label{fig:GEOexample}
\end{figure}
Different tissues, from neurons to blood, share the same genome but differ in gene expression.  Annotating gene expression data with tissue type is critical to enable data reuse for cell-development and cancer studies~\cite{rung2013reuse}. 
Lee et al. manually annotated a large dataset of 14,510 expression samples to train a state-of-the-art supervised classifier \cite{Lee2013}.
However, their dataset only covers 176 tissue types, or less than 4\% of classes in BRENDA Tissue Ontology.
In this section, we applied \EZL to learn a far more accurate classifier that can in principle cover all tissue types in BRENDA. (In practice, the coverage is limited by the available unlabeled gene expression samples; in our experiments \EZL learned to predict 601 tissue types.)

\smallpar{Annotation task}
The goal is to annotate gene expression samples with their tissue type.
The input is a gene expression profile (a 20,000-dimension vector with a numeric value signifying the expression level for each gene). The output is a tissue type.
We used the standard BRENDA Tissue Ontology \cite{Gremse01012011}, which contains 4931 human tissue types.
For gene expression data, we used the Gene Expression Omnibus (GEO)~\cite{edgar2002gene}, a popular repository run by the National Center for Biotechnology Information. 
Figure~\ref{fig:GEOexample} shows an example gene expression profile with text description in GEO.
We focused on the most common data-generation platform (Affymetrix U133 Plus 2.0), and obtained a dataset of 116,895 human samples.
Each sample was processed using Universal exPression Codes (UPC) \cite{Piccolo29102013} to minimize batch effects and normalize the expression values to [0,1]. 
Text descriptions were obtained from GEOmetadb \cite{Zhu01122008}.

\smallpar{Main classifier} 
We implemented $\tt Train_{main}$ using a deep denoising auto-encoder (DAE) with three LeakyReLU layers to convert the gene expression profile to a 128-dimensional vector \cite{vincent2008extracting}, followed by multinomial logistic regression, trained end-to-end in Keras \cite{Keras}, using L2 regularization with weight $1e-4$ and RMSProp optimizer \cite{tieleman2012lecture}.

\smallpar{Auxiliary classifier} 
We implemented $\tt Train_{aux}$ using fastText with their recommended parameters (25 epochs and starting learning rate of 1.0) \cite{fastText}. 
In principle, we can continue the alternating training steps until neither classifier's predictions change significantly. In practice, the algorithm converges quickly \cite{nigam2000analyzing}, and we simply ran all experiments with five iterations.

\smallpar{Systems}
We compared \EZL with URSA \cite{Lee2013}, the state-of-the-art supervised method that was trained on a large labeled dataset of 14,510 examples and used a sophisticated Bayesian method to refine SVM classification based on the tissue ontology. 
We also compared it with co-training \cite{blum1998combining} and co-EM \cite{nigam2000analyzing}, two representative methods for leveraging unlabeled data that also use an auxiliary view to support the main classification. Unlike \EZL, they require labeled data to train their initial classifiers. 
After the first iteration, high-confidence predictions on the unlabeled data are added to the labeled examples. In co-training, once a unlabeled sample is added to the labeled set, it is not reconsidered again, whereas in co-EM, all of them are re-annotated in each iteration. We found that co-training and co-EM performed similarly, so we only report the co-EM results.

\smallpar{Evaluation}
The BRENDA Tissue Ontology is a directed acyclic graph (DAG), with nodes being tissue types and directed edges pointing from a parent tissue to a child, such as $\tt leukocyte$ $\rightarrow$ $\tt leukemia\; cell$.
We evaluated the classification results using {\it ontology-based precision and recall}.   
We expand each singleton class (predicted or true) into a set that includes all ancestors except the root.
We then measure precision and recall as usual: precision is the proportion of correct predicted classes among all predicted classes, and recall is the proportion of correct predicted classes among true classes, with ancestors included in all cases. This metric closely resembles the approach by Verspoor et al. 
\cite{verspoor2006categorization}, except that we are using the ``micro'' version (i.e., the predictions for all samples are first combined before measuring precision and recall).
If the system predicts an irrelevant class in a different branch under the root, the intersection of the predicted and true sets is empty and the penalty is severe. If the predicted class is an ancestor (more general) or a descendent (more specific), the intersection is non-empty and the penalty is less severe, but overly general or overly specific predictions are penalized more than close neighbors.
We tested on the Comprehensive Map of Human Gene Expression (CMHGP), the largest expression dataset with manual tissue annotations \cite{Torrente2016}. CMHPG used tissue types from the Experimental Factor Ontology (EFO) \cite{malone2010modeling}, which can be mapped to the BRENDA Tissue Ontology. 
To make the comparison fair, 7,209 CMHGP samples that were in the supervised training set for URSA were excluded from the test set. The final test set contains 15,129 samples of 628 tissue types.

\begin{figure}
  \centering
  \includegraphics[width=0.8\linewidth]{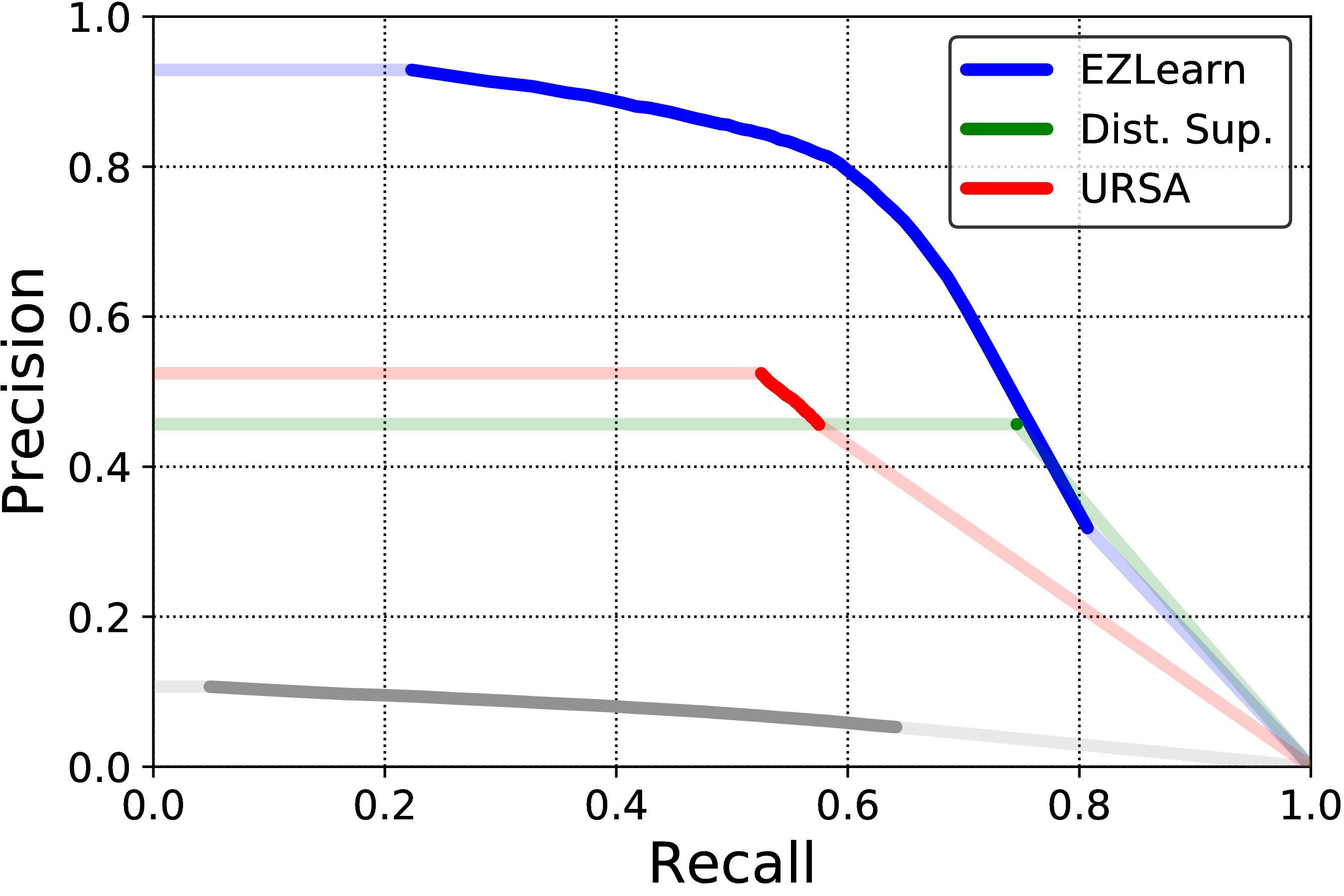}
  \vspace{-8pt}
  \caption{Ontology-based precision-recall curves comparing \EZL, distant supervision, URSA, and the random baseline (gray). Extrapolated points are shown in transparent colors.
    }
  \label{fig:geo_PR}
\end{figure}

\begin{figure}[!htb]
\begin{subfigure}{.47\textwidth}
\centering
  \includegraphics[width=0.8\linewidth]{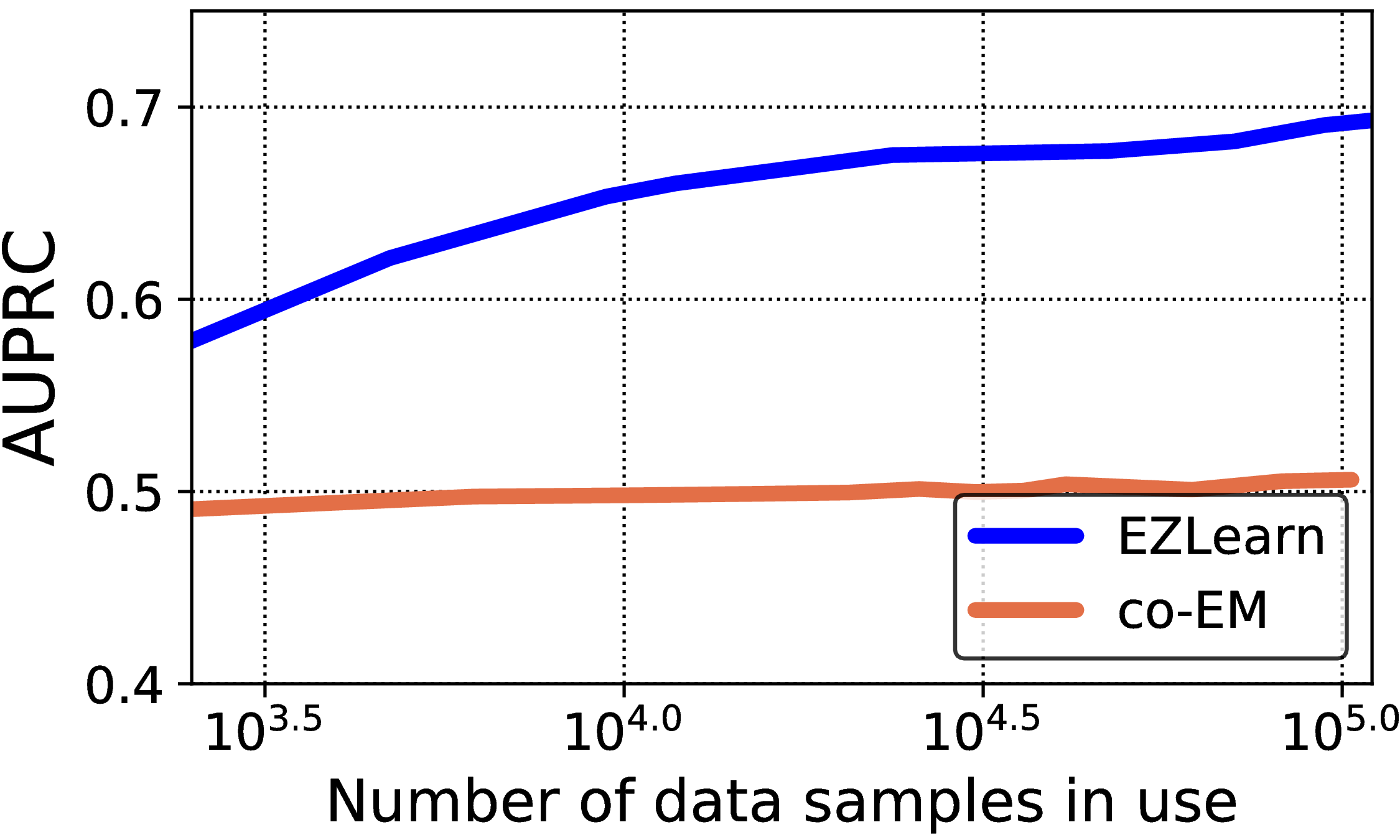}
  \caption{}
  \end{subfigure}\hfill
\begin{subfigure}{.47\textwidth}
\centering
  \includegraphics[width=0.8\linewidth]{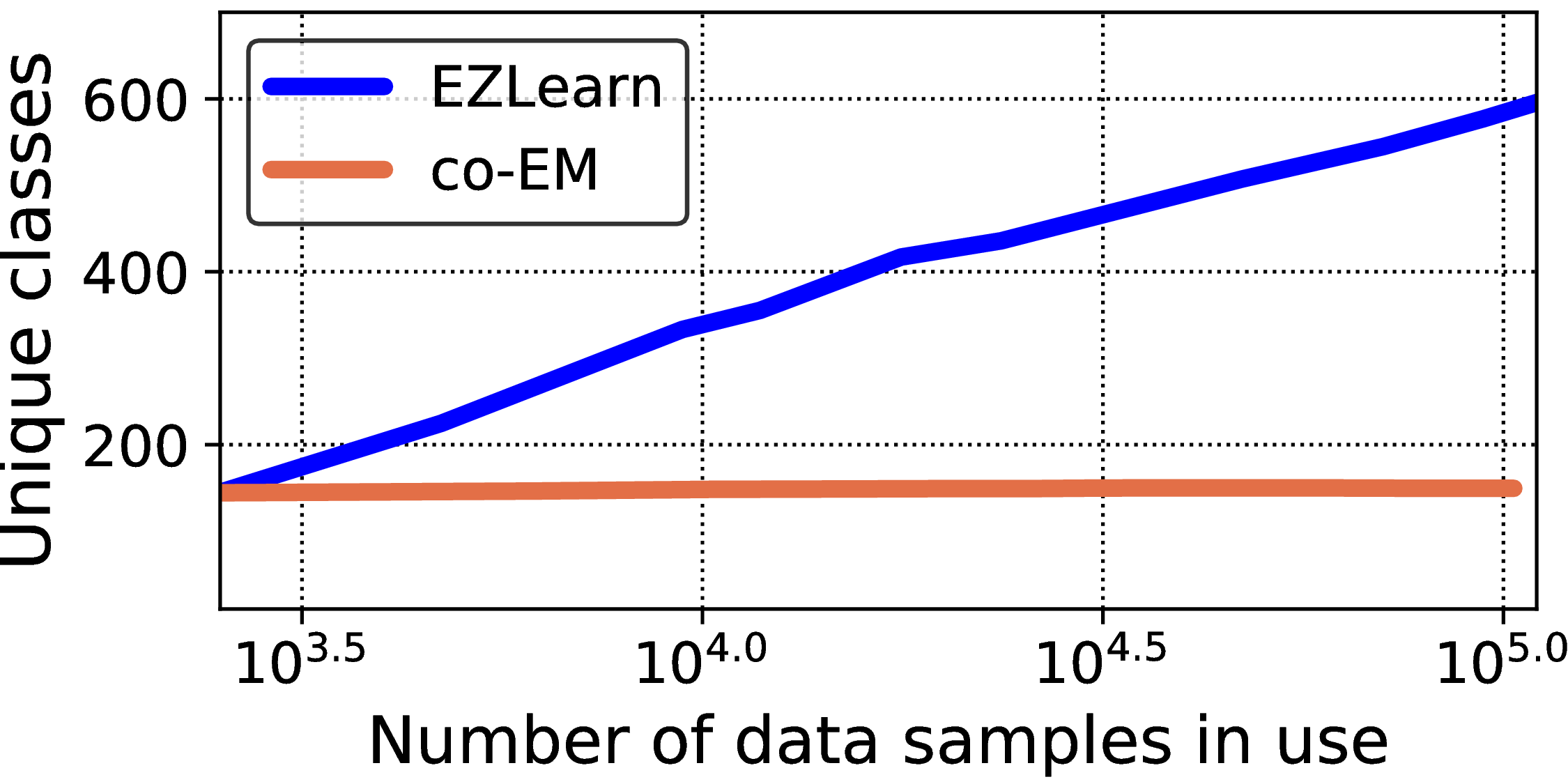}
  \caption{}
  \end{subfigure}
\vspace{-8pt}
\caption{
(a) Comparison of test accuracy with varying amount of unlabeled data, averaged over fifteen runs. \EZL gained substantially with more data, whereas co-EM barely improves. 
(b) Comparison of number of unique classes in high-confidence predictions with varying amount of unlabeled data. \EZL's  gain stems in large part from learning to annotate an increasing number of classes, by using organic supervision to generate noisy examples, whereas co-EM is confined to classes in its labeled data.
}
\label{fig:geo-figs}
\end{figure}

\smallpar{Results} 
We report both the area under the precision-recall curve (AUPRC) and the precision at 0.5 recall. Table~\ref{fig:genomicsTable} shows the main classification results (with $\tt Resolve=Relation$ in \EZL). 
Remarkably, without using any manually labeled data, \EZL outperformed the state-of-the-art supervised method by a wide margin, improving AUPRC by an absolute 27 points over URSA, and over 30 points in precision at 0.5 recall.
Compared to co-EM, \EZL improves AUPRC by 18 points and precision at 0.5 recall by 25 points. 
Figure~\ref{fig:geo_PR} shows the precision-recall curves.

To investigate why \EZL attained such a clear advantage even against co-EM, which used both labeled and unlabeled data and jointly trained an auxiliary text classifier, we compared their performance using varying amount of unlabeled data (averaged over fifteen runs).
Figure~\ref{fig:geo-figs}(a) shows the results. Note that the x-axis (number of unlabeled examples in use) is in log-scale. Co-EM barely improves with more unlabeled data, whereas \EZL improves substantially from 2\% to 100\% of unlabeled data.

To understand why this is the case, we further compare the number of unique classes predicted by the two methods. See Figure~\ref{fig:geo-figs}(b).
Co-EM is confined to the classes in its labeled data and its use of unlabeled data is limited to the extent of improving predictions for those classes.
In contrast, by using organic supervision from the lexicon and text descriptions, \EZL can expand the classes in its purview with more unlabeled data, in addition to improving predictive accuracy for individual classes.
The gain seems to gradually taper off (Figure~\ref{fig:geo-figs}(a)), but we suspect that this is an artifact of the current test set. Although CMHGP is large, the number of tissue types in it (628) is still a fraction of that in the BRENDA Tissue Ontology (4931). 
Indeed, Figure~\ref{fig:geo-figs}(b) shows that the number of its predicted classes keeps climbing. This suggests that with additional unlabeled data \EZL can improve even further, and with additional test classes, the advantage of \EZL might become even larger.

We also evaluated on the subset of CMGHP with tissue types confined to those in the labeled data used by URSA and co-EM, to perfectly match their training conditions.
Unsurprisingly, URSA and co-EM performed much better, attaining 0.53 and 0.67 in AUPRC, respectively (though URSA's accuracy is significantly lower than its training accuracy, suggesting overfitting).
Remarkably, by exploiting organic supervision, \EZL still outperformed both URSA and co-EM, attaining 0.71 in AUPROC in this setting.

\begin{figure}[t!]
\centering
  \includegraphics[width=0.75\linewidth]{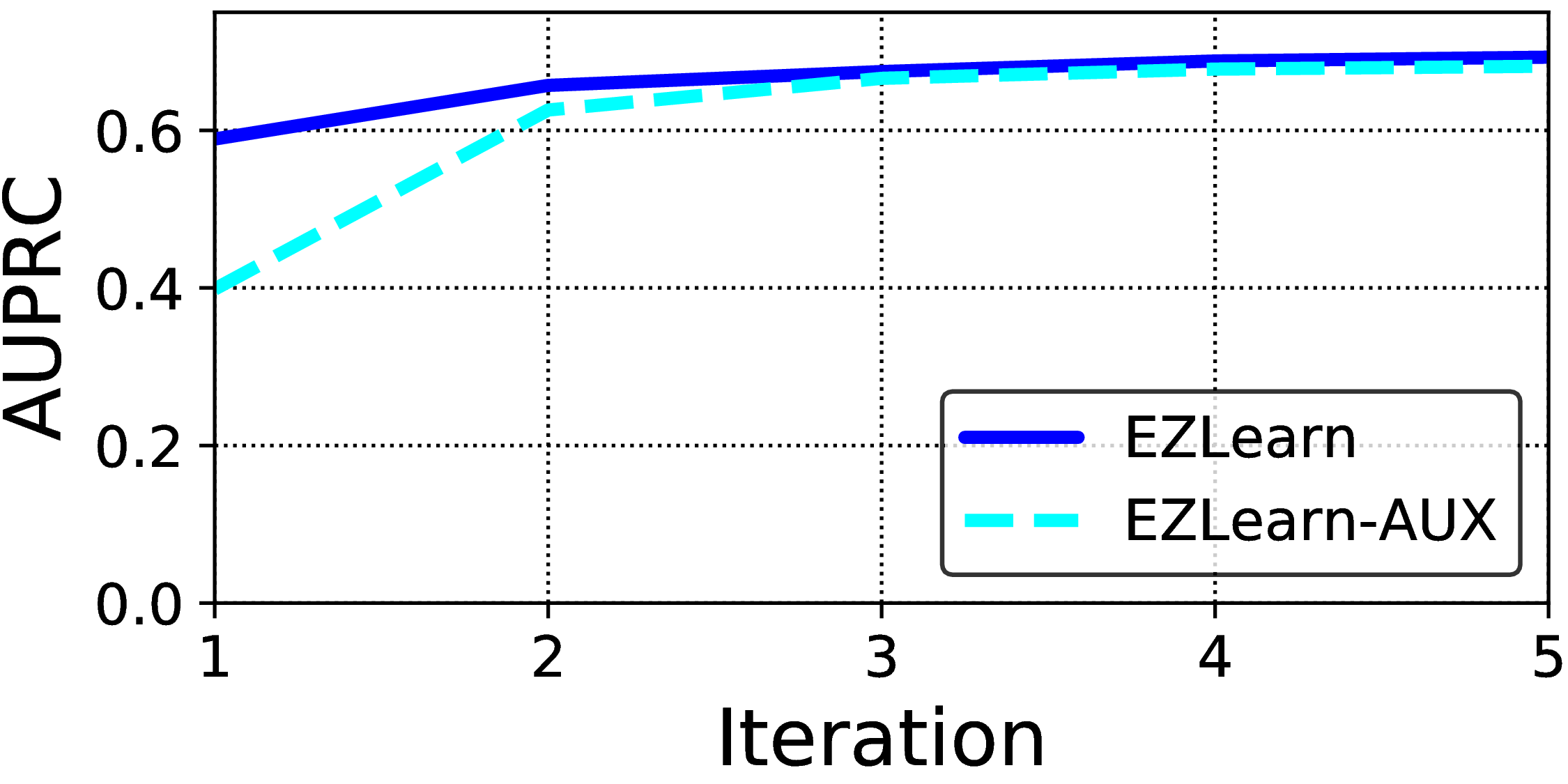}
  \caption{Comparison of test accuracy of the main and auxiliary classifiers at various iterations during learning.}
\label{fig:geo-iters-figure}
\end{figure}

\EZL amounts to initializing with distant supervision (first iteration) and continuing with an EM-like process as in co-training and co-EM.
This enables the main classifier and the auxiliary text classifier to improve each other during learning (Figure~\ref{fig:geo-iters-figure}).
Overall, compared to distant supervision, adding co-training led to further significant gains of 10 points in AUPRC and 23 points in precision at 0.5 recall (Table 1).

If labeled examples are available, \EZL can simply add them to the labeled sets at each iteration.
After incorporating the URSA labeled examples \cite{Lee2013}, the AUPRC of \EZL improved by two absolute points, with  precision at 0.5 recall increasing to 0.87 (not shown in Table 1).

\begin{table}[]
\centering
{\small
\begin{tabular}{|c|c|c|c|c|c|}
\hline
$\tt Resolve$ & $\tt Stand.$ & $\tt Pred.$ & $\tt Union$ & $\tt Inter.$ & $\tt Relat.$ \\ \hline
\textbf{\# Classes} & 623 & 329 & 603 & 351 & 601 \\ \hline
\textbf{AUPRC} & 0.59 & 0.64 & 0.59 & 0.66 & {\bf 0.69} \\
\hline
\end{tabular}
}
\vspace{-2pt}
\caption{Comparison of test results and numbers of unique classes in high-confidence predictions on the Comprehensive Map of Human Gene Expression by \EZL with various strategies in resolving conflicts between distant supervision and classifier prediction.}
\label{tbl:DistSupervisionResults}
\end{table}

\begin{figure}[]
  \centering
  \includegraphics[width=0.75\linewidth]{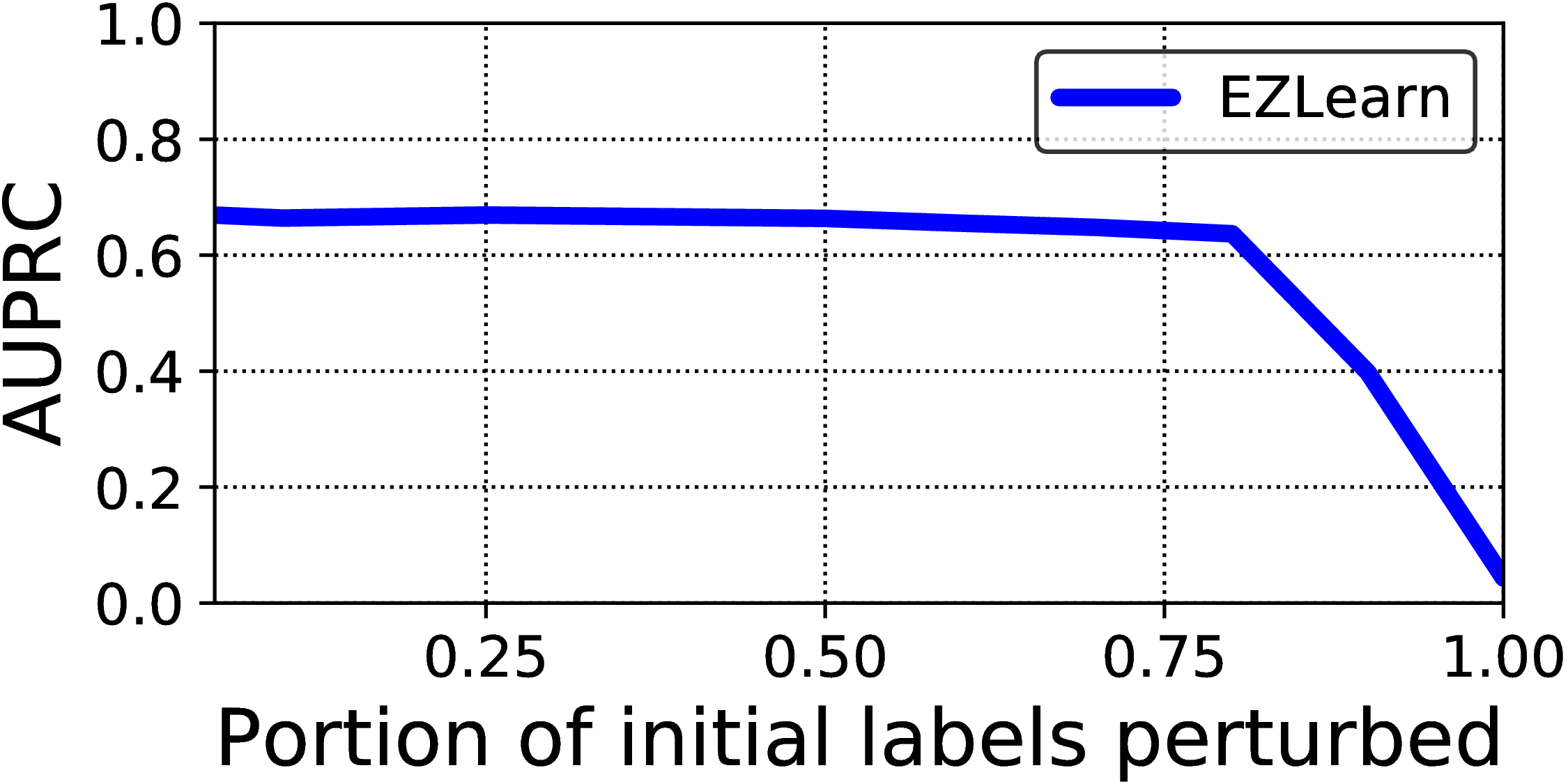}
  \caption{\EZL's test accuracy with varying portion of the distant-supervision labels replaced by random ones in the first iteration. \EZL is remarkably robust to noise, with its accuracy only starting to deteriorate significantly after 80\% of labels are perturbed.
    }
  \label{fig:geo_label_shuffle}
\end{figure}

Compared to direct supervision, organic supervision is inherently noisy. 
Consequently, it is generally beneficial to reconcile classifier prediction with distant supervision when they are in conflict, as Table~\ref{tbl:DistSupervisionResults} shows. $\tt Standard$ (always choosing distant supervision when available) significantly trailed the alternative approach that always picks classifier's prediction ($\tt Predict$). $\tt Union$ predicted more classes than $\tt Intersect$ but suffered large precision loss. By taking into account of hierarchical relations in the class ontology, $\tt Relation$ substantially outperformed all other methods in accuracy, while also covering a large number of classes.

To evaluate \EZL's robustness, we simulated noise by replacing a portion of the initial distant-supervision labels with random ones.
Figure~\ref{fig:geo_label_shuffle} shows the results. 
Interestingly, \EZL can withstand a significant amount of label perturbation: test performance only deteriorates drastically  when more than 80\% of initial labels are replaced by random ones. This result suggests that \EZL can still perform well for applications with far more noise in their organic supervision.

\vspace{-5pt}
\section{Application: Figure Comprehension}

\begin{figure}
\centering
            \includegraphics[width=0.85\linewidth]{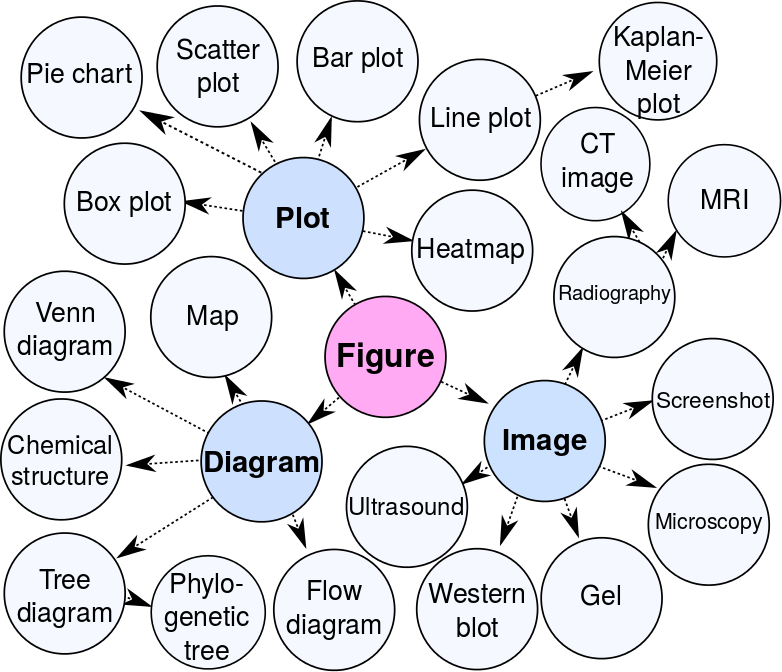}
    \vspace{-4pt}
  \caption{The Viziometrics project only considers three coarse classes $\tt Plot$, $\tt Diagram$, and $\tt Image$ for figures due to high labeling cost. We expanded them into 24 classes, which \EZL learned to accurately predict with zero manually labeled examples.
    }
\label{fig:vizio-ontology}
\end{figure}

Figures in scientific papers communicate key results and provide visual explanations of complex concepts.
However, while text understanding has been intensely studied, 
figures have received much less attention in the past.
A notable exception is the Viziometrics project \cite{lee2017viziometrics}, which annotated a large number of examples for classifying scientific figures.
Due to the considerable cost of labeling examples, they only used five coarse classes: $\tt Plot$, $\tt Diagram$, $\tt Image$, $\tt Table$ and $\tt Equation$. We exclude the last two as they do not represent true figures.
In practice, figure-comprehension projects would be much more useful if they include larger set of specialized figure types. 
To explore this direction, we devised an ontology where $\tt Plot$, $\tt Diagram$, and $\tt Image$ are further refined into a total of twenty-four classes, such as $\tt Box plot$, $\tt MRI$ and $\tt Pie Chart$ (Figure~\ref{fig:vizio-ontology}). \EZL naturally accommodates a large and dynamic ontology since no manually labeled data is required.

\smallpar{Annotation task}
The goal is to annotate figures with semantic types shown in Figure~\ref{fig:vizio-ontology}.
The input is the image of a figure with varying size.
The output is the semantic type.
We obtained the data from the Viziometrics project \cite{lee2017viziometrics} through its open API. For simplicity, we focused on the non-composite subset comprising single-pane figures, yielding 1,174,456
figures along with free-text captions for use as distant supervision. 
As in the gene expression case, captions might be empty or missing.

\smallpar{System}
Each figure image was first resized and converted to a 2048-dimensional real-valued vector using a convolutional neural network \cite{he2016deep} trained on ImageNet \cite{deng2009imagenet}.
We follow \cite{howe2017deep} and use the ResNet-50 model with pre-trained weights provided by Keras \cite{Keras}.
We used the same classifiers and hyperparameters as in the functional genomics application.
We used a lexicon that simply comprises of the names of the new classes, and compared \EZL with the Viziometrics classifier.
We also compared with a lexicon-informed baseline that annotates a figure with the most specific class whose name is mentioned in the caption (or root otherwise).

\begin{figure}
    \centering
    \includegraphics[width=0.85\linewidth]{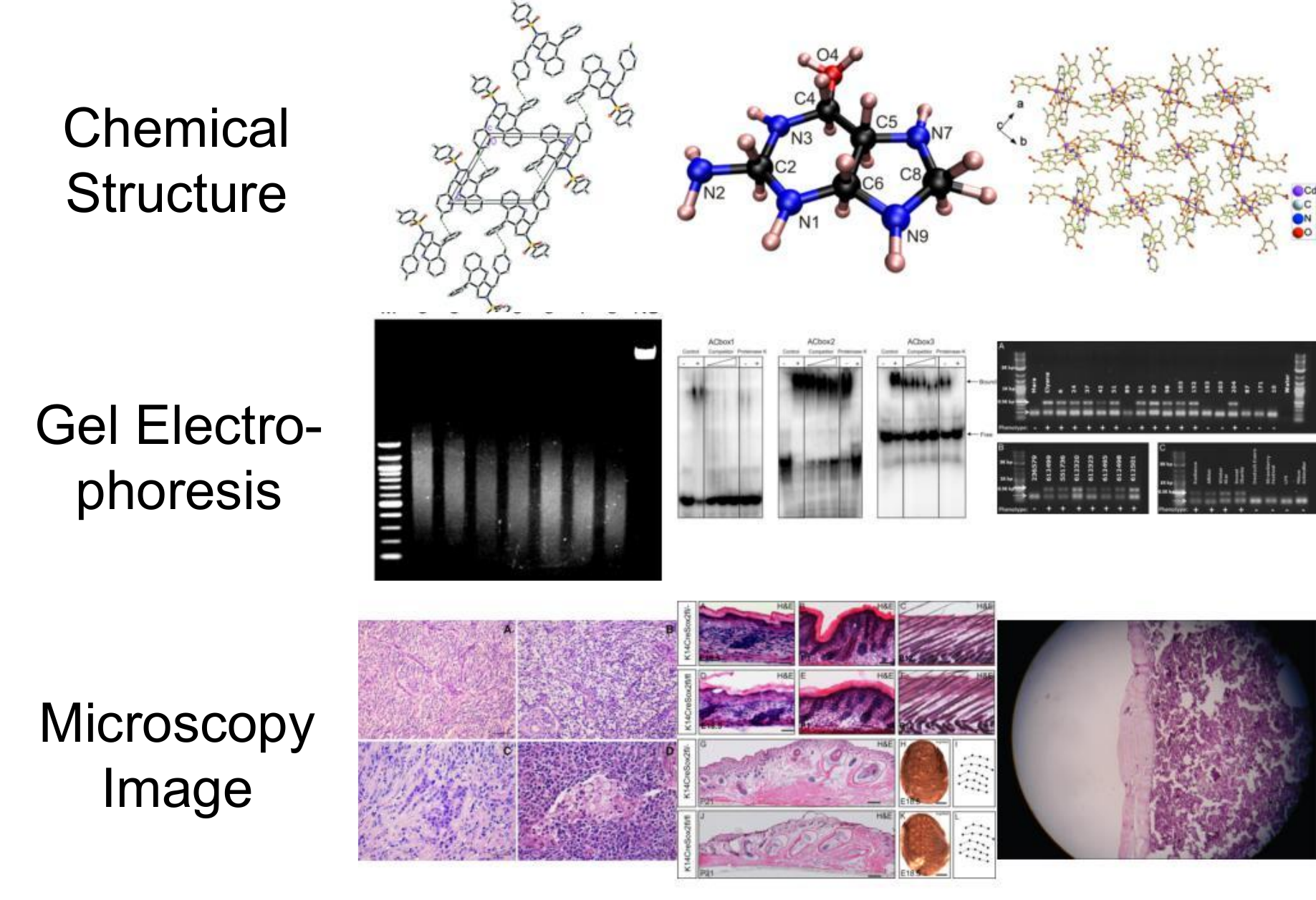}
    \caption{Example annotations by \EZL, all chosen among figures with no class information in their captions.}
\label{fig:vizio-results}
\end{figure}

\smallpar{Evaluation}
We followed the functional genomics application and evaluated on ontology-based precision and recall. 
Since the new classes are direct refinement of the old ones, we can also evaluate the Viziometrics classifier using this metric.
To the best of our knowledge, there is no prior dataset or evaluation for figure annotation with fine-grained semantic classes as in Figure~\ref{fig:vizio-ontology}. Therefore, we manually annotated an independent test set of 500 examples.

\begin{table}[!ht]
\centering

\begin{tabular}{|l|c|c|c|c|}
\hline
\textbf{} & Lexicon & Vizio. & Dist. Sup. & \EZL \\ \hline
AUPRC    & 0.44     & 0.53   & 0.75 & $\mathbf{0.79}$     \\ \hline
Prec@0.5 & 0.31     &  0.43  & 0.87 & $\mathbf{0.92}$ \\ \hline
\end{tabular}
\vspace{-4pt}
\caption{Comparison of test results between \EZL, the lexicon baseline, the Viziometrics classifier, and distant supervision.}
\label{fig:vizioTable}
\end{table}

\vspace{-10pt}
\smallpar{Results}
\EZL substantially outperformed both the lexicon-informed baseline and the Viziometrics classifier (Table~\ref{fig:vizioTable}).
The state-of-the-art Viziometrics classifier was trained on 3271 labeled examples, and attained an accuracy of 92\% on the coarse classes. 
So the gain attained by \EZL reflects its ability to extract a large amount of fine-grained semantic information missing in the coarse classes.
Figure~\ref{fig:vizio-results} shows example annotations by \EZL, all chosen from figures with no class mention in the caption. 

 \vspace{-4pt}
\section{Conclusion}

We propose \EZL for automated data annotation, by combining distant supervision and co-training.
\EZL is well suited to high-value domains with numerous classes and frequent update.
Experiments in functional genomics and scientific figure comprehension show that \EZL is broadly applicable, robust to noise, and capable of learning accurate classifier without manually labeled data, even outperforming state-of-the-art supervised systems by a wide margin.

\small

\bibliography{references}
\bibliographystyle{named}
\end{document}